\documentclass{article} 
\usepackage{iclr2019_conference,times}

\usepackage{amsmath,amsfonts,bm}

\def\eqref#1{equation~\ref{#1}}

\def\1{\bm{1}}

\DeclareMathAlphabet{\mathsfit}{\encodingdefault}{\sfdefault}{m}{sl}
\SetMathAlphabet{\mathsfit}{bold}{\encodingdefault}{\sfdefault}{bx}{n}

\usepackage[utf8]{inputenc} 
\usepackage[T1]{fontenc}    
\usepackage{hyperref}       
\usepackage{url}            
\usepackage{booktabs}       
\usepackage{amsfonts}       
\usepackage{nicefrac}       
\usepackage{microtype}      
\usepackage{amsmath}         
\usepackage{pgfplots}
\usepackage{cleveref}        
\usepackage{booktabs}        
\usepackage{nicefrac}        
\usepackage{tikz}
\usepackage{pgfplots}
\usepackage{graphicx}        
\usepackage{subfig}
\usepackage[toc,page]{appendix}
\usepackage[final]{pdfpages}

\DeclareMathOperator{\tr}{tr}

\newcommand{\mb}[1]{\mathbf{#1}}
\newcommand{\todo}[1]{}

\title{Static activation function normalization}

\author{
  Pierre H. Richemond \\
  Data Science Institute \\
  Imperial College London \\
  London, SW7 2AZ \\
  \texttt{phr17@imperial.ac.uk} \\
  \and
  Yike Guo \\
  Data Science Institute \\
  Imperial College London \\
  London, SW7 2AZ \\
  \texttt{y.guo@imperial.ac.uk} \\
}

\iclrfinalcopy 
\begin{document}

\maketitle

\begin{abstract}
 Recent seminal work at the intersection of deep neural networks practice and random matrix theory has linked the convergence speed and robustness of these networks with the combination of random weight initialization and nonlinear activation function in use. Building on those principles, we introduce a process to transform an existing activation function into another one with better properties. We term such transform \emph{static activation normalization}. More specifically we focus on this normalization applied to the ReLU unit, and show empirically that it significantly promotes convergence robustness, maximum training depth, and anytime performance. We verify these claims by examining empirical eigenvalue distributions of networks trained with those activations. Our static activation normalization provides a first step towards giving benefits similar in spirit to schemes like batch normalization, but without computational cost.
\end{abstract}

\section{Introduction}

Training very deep networks via stochastic gradient descent is an unstable numerical procedure. The end result of that training varies in quality based on metrics such as convergence speed, robustness to random factors, and of course terminal accuracy. Various moving parts of the deep network in question can be interchanged in order to favour one aspect over the other. Of particular interest is the maximum training depth, or number of layers, that a specific architecture can reliably achieve. A key breakthrough in deep learning was achieved when that number got in the region of 10 layers by switching the nonlinear activation unit from traditionally used hyperbolic tangent and sigmoid units to a rectified linear unit \citep{xavier}. Further progress, by an order of magnitude or two, was recently made possible by the combination of a residual network reparameterization \citep{heresidual} and a milestone process called batch normalization \citep{batchnorm}, constraining the output data statistics after each layer. For all intents and purposes, these two methods have ushered in an era of arbitrarily deep networks, at the expense of increased computational cost (and maybe less theoretical interpretability). Without using them the number of layers that can be trained reliably is still limited \citep{pennington2017resurrecting}. For instance, we did not manage to train feedforward networks deeper than 30 layers, using rectified linear units, without such stabilising procedures. In this work, we investigate if existing theory available for multi-layer perceptrons enables us to design activation functions more suitable for training deeper networks, by transforming existing activations.

The contributions of this paper are as follows :
\begin{itemize}
\item first, we use newly discovered applications of random matrix theory to neural networks in order to give a new, general process that transforms activation functions;
\item second, we compute the transforms of activation functions commonly used in deep learning, then give a closed-form expression for the transformed rectified linear unit, and focus on evaluating it empirically;
\item third, we explore the empirical eigenvalue spectra of networks trained with these new activations.
\end{itemize}

\section{Related Work}

Studies in physics-inspired theoretical properties of the gradient flow in neural networks go, for instance, as far back as Sompolinsky \citep{sompolinsky}. A revival of this literature was sparked by the Stanford school of thought \citep{exponential, deepinfo}, making connections between neural networks and statistical physics, in particular noting the existence of a critical phase separating the two unstable regimes of exploding and vanishing gradients.

On the activation function front, sigmoidal and hyperbolic tangent units have historically been favoured with neural networks. The advent of rectified linear units, or ReLUs \citep{xavier, nair2010rectified} marked a notable shift that enabled the training of deeper networks than previously thought possible. This sparkled much research aimed at maximizing accuracy, with  exponential linear units or ELUs \citep{clevert2015fast, klambauer2017self}, Gaussian linear units or GELUs \citep{hendrycks2016bridging}, and \emph{swish} functions \cite{elfwing2018sigmoid} all notable units performance-wise, enabling accuracy gains.

The random initialization of the weights of a neural network plays an important role. Historically, Glorot \citep{xavier} and then He \citep{he2015delving} both considered schemes adapted to a Gaussian distribution of weights, computing relevant variances ; whereas Saxe \citep{saxe2013} investigated the necessity and benefits of initializing with weight matrices satisfying an orthogonality condition. Until recently the choices of weight initialization and activation function were made mostly separately, but recent work has elucidated that they are both related by complex interplay \citep{deepuniversality}. Such an advance in theoretical understanding was made possible by the use of random matrix theory (\citep{pennington, pennington2017resurrecting, penningtonnonlinear, couillet} ) in the large square layer regime for feed-forward neural networks. It opens up the possibility to train stabler, deeper networks without requiring techniques like skip connections \citep{penningtonnonlinear}, batch normalization, \citep{batchnorm} and its siblings weight normalization \citep{weightnorm} or group normalization \citep{groupnorm}. A method that would 'push the batch norm inside the activations', at least in a static manner, would be of computational interest.

\section{Review of random matrices and free probability for neural networks}

\subsection{Dynamical isometry}

For clarity, we begin with laying out our theoretical framework for what follows. This expository section is much indebted to the work of Pennington \citep{deepuniversality, penningtonnonlinear}, and the reader familiar with his work might omit it. Let us assume we have an $L$-layer deep multi-layer perceptron, or feed-forward neural network, of width $N$, with weight matrices $\mb{W}_l\in \mathbb{R}^{N\times N}$,  bias vectors $\mb{b}_l$, pre-activations $\mb{h}_l$, and post-activations $\mb{x}_l$, for each layer $l$ in $1,\dots,L$.  The forward propagation equations are
\begin{equation}
\label{eqn:dynam}
\mb{x}_l = \phi(\mb{h}_l)\,,\quad \mb{h}_l = \mb{W}_l \mb{x}_{l-1} + \mb{b}_l\,,
\end{equation}
where $\phi : \mathbb{R} \to \mathbb{R}$ is a nonlinearity (also referred to in this paper as an \emph{activation function}), and the input is  $\mb{x}_0 \in \mathbb{R}^N$.  The input-output Jacobian $\mb{J} \in \mathbb{R}^{N\times N}$ is therefore given by
\begin{equation}
\begin{split}
\label{eqn:Jz}
\mb{J} = \frac{\partial \mb{x}_L}{\partial \mb{x}_0} = \prod_{l=1}^L \mb{D}_l \mb{W}_l.
\end{split}
\end{equation}
Here $\mb{D}_l$ is a diagonal matrix with diagonal terms $D_{l,ij} = \phi'(h_{l,i}) \delta_{ij}$, where $\delta_{ij}$ is a Kronecker delta (it will in general denote a Dirac delta in the sequel). Intuitively, this input-output Jacobian $\mb{J}$, and particularly its spectral content - eigenvalues and conditioning - is critical for the operation of backpropagation that happens during the training of the network : if we knew that \emph{all} eigenvalue moduli were tightly concentrated around a mean value of 1, gradient flow in the network would be optimal, as no direction of input data would yield neither vanishing or exploding gradients. This very desirable situation is called dynamical isometry. This way of normalizing networks by construction ensures network training proceeds as fast as possible.
The biases don't come into play in the expression of $\mb{J}$, so we need consider weights only - as random matrices who belong to an ensemble that depends on how the neural network is \emph{initialized}. Now we want to study the whole spectral distribution of Jacobian $\mb{J}$ for $W_l \in \mathbb{R}^N$, in the wide layer limit $N \rightarrow \infty$. In general, given a random matrix $\mb{X}$, its limiting spectral density is a real function defined as the limit of the eigenvalue empirical measure
\begin{equation}
\rho_{X}(\lambda) \equiv \underset{N \rightarrow \infty}{\lim} \mathbb{E}_{X} \left[ \frac{1}{N} \sum_{i=1}^N \delta(\lambda - \lambda_i) \right] ,
\end{equation}
Since we only care about the squared absolute values of the eigenvalues, we can consider the limiting spectral density of the self-adjoint and therefore semidefinite positive $\mb{J} \mb{J}^T$, so as to compare $\rho_{J J^T}(\lambda)$ and the ideal distribution $\lambda \rightarrow \delta(\lambda)-1$. The Jacobian $\mb{J}$ is a product of random matrices as given by equation \ref{eqn:Jz}, so the mathematical problem we are facing is that of \emph{computing the eigenvalues of a product of random matrices}.

To this end we consider two complex variable transforms $M$ and $S$, that turn a random matrix ensemble's spectral density into a complex variable function. These transforms both encode the same information; we use them interchangeably as per ease of computation. The first one, $M_X$, is just the moment generating function,
\begin{equation}
\label{eqn:moments}
M_X(z) \equiv \sum_{k=1}^{\infty} \frac{m_k}{z^k}\,,
\end{equation}
where $m_k$ is the $k$-th moment of the distribution $\rho_X$, or the limit trace moment of random matrix $X$,
\begin{equation}
\label{eqn:traces}
m_k  = \int \lambda^k \rho_X(\lambda) d\lambda \; = \underset{N \rightarrow \infty}{\lim}\frac{1}{N} \mathbb{E}_X \left[ \tr \mb{X}^k \right]\,.
\end{equation}
This information that encodes all the $m_k$ coefficients is also contained in the so-called Cauchy-Stieltjes transform turning a random matrix ensemble into a function of the complex variable:
\begin{equation}
G_{X}(z) = - \mathbb{E}[\tr(X-zI)^{-1}]
\end{equation}

On the other hand, the \emph{S-transform} is defined as
\begin{equation}
\label{eqn:SMrelation}
S_X(z) = \frac{1+z}{z  M_X^{-1}(z)}\,.
\end{equation}
where $M_X^{-1}(z)$ is the functional inverse of $M_X(z)$.
The S-transform is key here, due to its morphism behavior with respect to matrix multiplication. If $\mb{A}$ and $\mb{B}$ are two freely independent\footnote{The theory of \emph{free independence} dates to Voiculescu, and relies on an assumption of commutatitivity of joint trace moments. This assumption holds more and more true in very high dimension where, due to the phenomenon of concentration of measure, random directions are orthogonal with high probability; and as such, eigenspaces of two random matrices are approximately orthogonal and approximately commute. Readers (should there be any) are referred to Speicher and Mingo \citep{mingo2017free} for more details.} random matrices, then the S-transform of the product random matrix ensemble is the product of their S-transforms :
\begin{equation}
\label{eqn:Stransform}
S_{AB}(z) = S_A(z) S_B(z)\,.
\end{equation}

This property then enables us to write in full for the self-adjoint Jacobian $\mb{J}\mb{J}^T$, and under an assumption of homogeneity across layers, that
\begin{equation}
S_{J J^T}(z) = {S_{D^2}^{L}(z) S_{W W^T}^{L}(z)}
\end{equation}
This gives us a strategy of computing the individual S-transforms of both activations and initializations, multiplying them together, and finally, inverting the result. However, attempting to write closed form S-transforms for all activation functions used in deep learning generally gives rise to intractable complex integrals.

\subsection{Static isometry : Nonlinearity M-Transform}

First, for any nonlinearity $f(x)$, we have, 
\begin{equation}
\label{eqn:MD2}
M_{D^2}(z) = \int \mathcal{D}h \frac{f'(\sqrt{q*} h)^2}{z - f'(\sqrt{q^*}h)^2}\,.
\end{equation}
The integral over the Gaussian measure $ \mathcal{D}h$ reflects a sum over all the activations $h^l_i$ in a layer, since in the large $N$ limit the empirical distribution of activations converges to a Gaussian with limit variance $q^*$ via central limit theorem. This complex parameter integral is in general intractable, but can be developed in power series involving 
\begin{equation}
M_{D^2}(z) \equiv  \sum_{k=1}^{\infty} \frac{\mu_k}{z^k} , \qquad \mu_k = \int \mathcal{D}h\; f'(\sqrt{q^*} h)^{2k}
\end{equation}

Based on this, Gaussian moments of the gradient'ed activation function are crucial controls for the mathematical properties of the activation function in question.

\section{Static activation normalization}\label{sec:normalization}

\subsection{Marcenko-Pastur static isometry}

Many of the Gaussian integrals in the $\mu_k$ above are \emph{de facto} intractable for numerous activation functions, even after approximations. However our main goal is to design activation functions as close as possible to isometry, static or dynamical.
For that, we introduce two measures of bias and variance shift respectively for a given activation $f$, that are
\begin{equation}
\xi = \left( \int{f'(z \cdot \sigma_w \sigma_x ) \mathcal{D}z} \right)^2
\end{equation}
and we want $\xi = 0$ for no bias shift (the gradient of the activation function, possibly dilated, is a zero Gaussian mean function). For the variance shift we want the second-order condition
\begin{equation}
\eta = \int{f(z \cdot \sigma_w \sigma_x )^2 \mathcal{D}z}
\end{equation}
where clearly $\eta = 1$ is desirable. 

Under those two conditions, Pennington \citep{penningtonnonlinear} shows that the Stieltjes transform $G_{M}(z)$ verifies the quadratic
\begin{equation}
z G_{M}^{2}(z) - G_{M}(z) + 1 = 0
\end{equation}
Solving and inverting it shows the eigenvalues distribution of $M = \frac{1}{m}f(W X)^t f(W X)$, and $X^t X$ are the same: they both follow a Marcenko-Pastur distribution.
Intuitively, this comes from the fact $X$ is assumed to be a Gaussian datapoint vector, so that $X^t X$ is a \emph{Wishart} random matrix, and the aforementioned conditions ensure that static spectral isometry is conserved. Numerical evidence also shows that this result holds too independently of the Gaussian character of $X$. This is a very desirable property termed \emph{static isometry} : when it holds, pairwise, fully-connected layers of a neural network maintain a constant (in modulus) eigenvalue distribution. In particular, this means that all higher moments of the data distribution, and its conditioning, get preserved at the beginning of training.

\begin{figure}[h]
\includegraphics[width=0.45\textwidth]{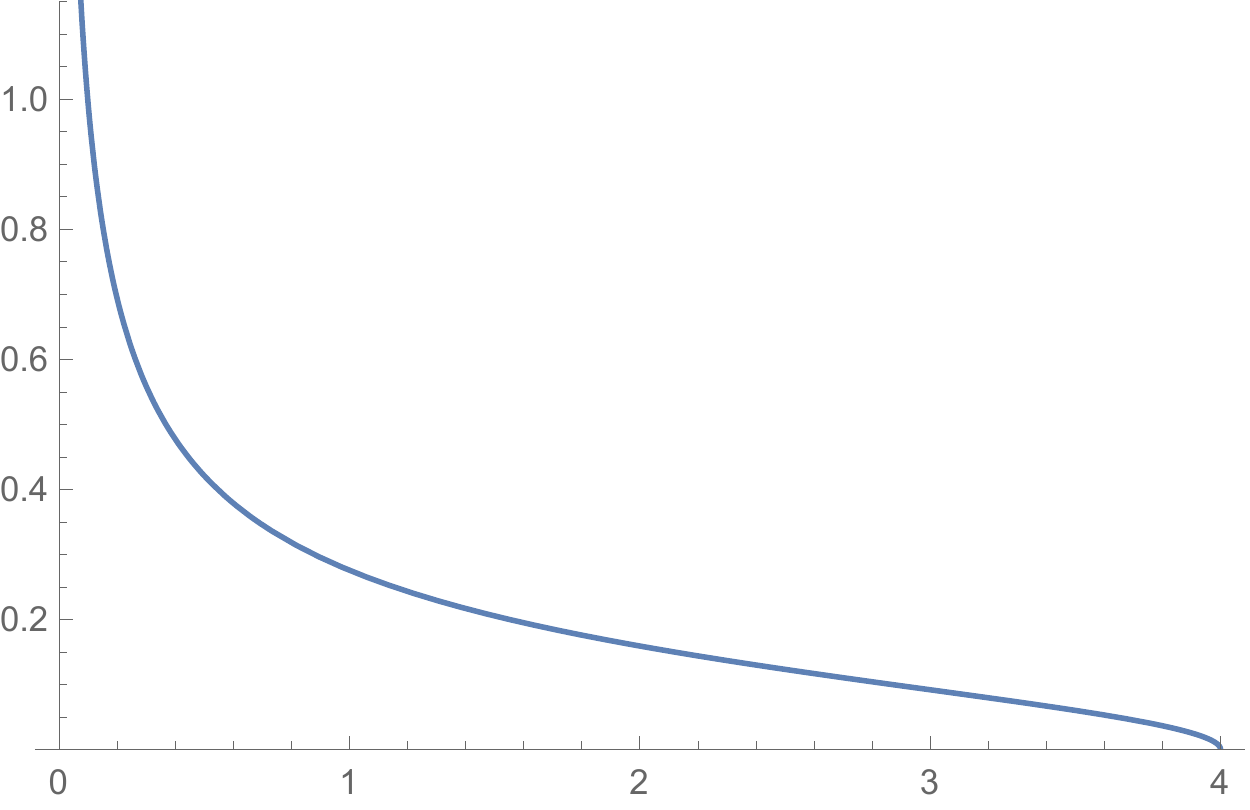}
\includegraphics[width=0.55\textwidth]{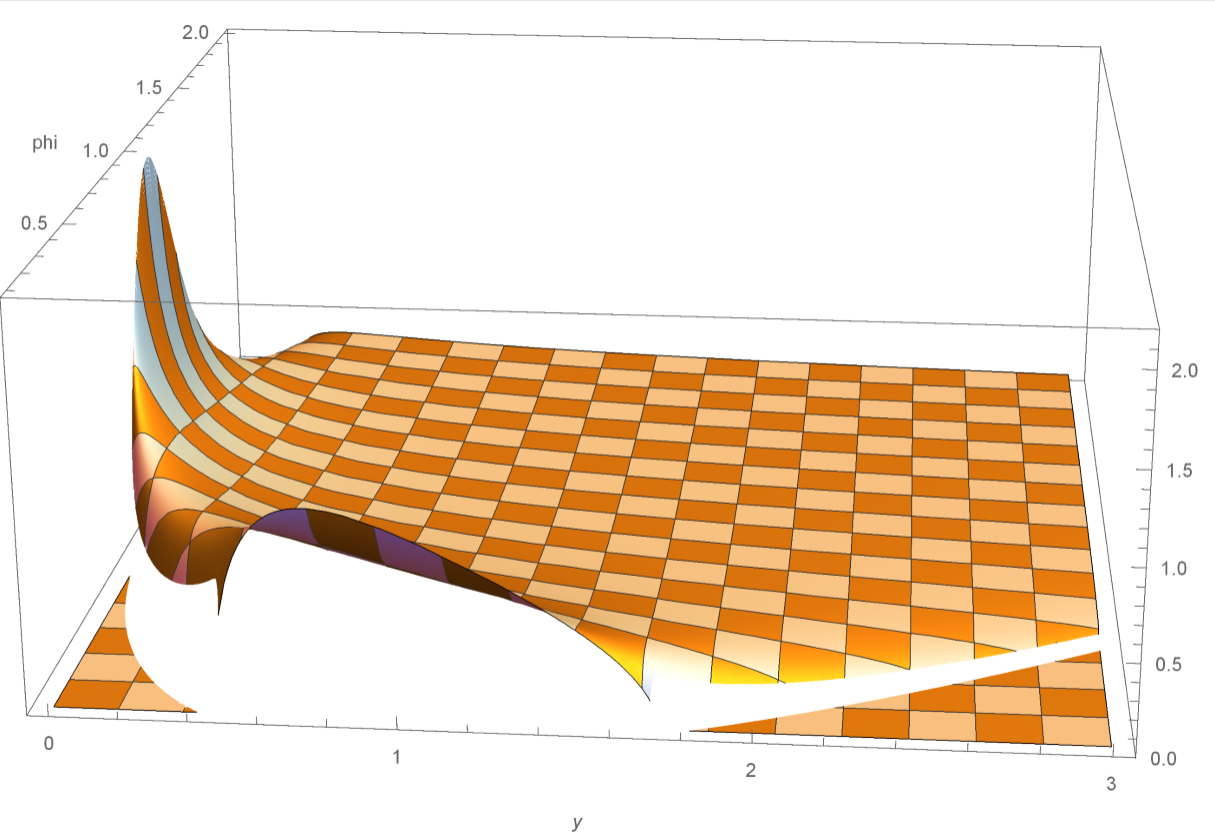}
\caption{A view of the Marcenko-Pastur distribution, for several values of the shape parameter $\phi$: on the left sliced as $\phi=1$, on the right stylized and represented with $\phi$ as the depth axis. Best viewed in colour.}
\label{fig:figure1}
\end{figure}

The probability density of the Marcenko-Pastur distribution, depicted figure \ref{fig:figure1}, with parameter $\phi \leq 1$ is given by :
\begin{equation}
\frac{1}{2\pi \phi} \frac{\sqrt{((1+\sqrt{\phi})^2-x)(x-(1-\sqrt{\phi})^2)}}{x} dx
\end{equation}

$\phi$ is the rectangularity ratio of neural network layers, which we pick square so that it is equal to 1. For all that follows, we will also further assume that we are in the limit of large square neural network layers, and that we normalize $\sigma_w$ and $\sigma_x$ to both be equal to $1$; this can be achieved by the choice of weight initialization for the network, and by accordingly pre-processing the dataset, respectively.

\subsection{Hermite polynomial basis expansion}

We can now hope to intervene on desirable properties of an activation function $f$ via its Gaussian moments. Those in turn come into play in the \emph{Hermite polynomial} series expansion of the activation function, as Hermite polynomials are orthogonal with respect to the standard Gaussian measure, as follows:
\begin{equation}
f(x) = \sum_{n=0}^{\infty}{\frac{f_n}{\sqrt{n!}} H_n(x)}
\end{equation}
where the Hermite orthogonal polynomials $H_n$ are defined by
\begin{equation}
H_{n}(x) = (-1)^n e^{\frac{x^2}{2}} \frac{\partial}{\partial x^n} e^{\frac{-x^2}{2}}
\end{equation}
Trivially $\xi = f_{1}^2$ and $H_1(x) = x$, so that $f(x) = f_0 + x \sqrt{\xi} + \sum_{n=2}^{\infty}{\frac{f_n}{\sqrt{n!}} H_n(x)}$. 
Further, $\eta - \xi = \sum_{n=1}^{\infty}{f_{n}^2}$.

\subsection{Static normalization : Principle}
Most activation functions do not satisfy these two conditions by default. However, if we make the simplifying assumption that $\sigma_x = \sigma_w \approx 1$ (that is, the network is initialized close to critical variance of the weights, and inputs have been suitably normalized too), then it is not hard to see that we can \emph{normalize} any activation function $f(x)$ statically by the transform below, scaling and removing the affine part:
\begin{equation}\label{san}
f(x) \leftarrow \frac{f(x) - x \int{f'(z) \mathcal{D}z} - \int{f(z) \mathcal{D}z} }{\sqrt{ \int{\left( f(x) - x \int{f'(z) \mathcal{D}z - \int{f(z) \mathcal{D}z}} \right)^2 \mathcal{D}x} }}
\end{equation}

which will roughly enforce Marcenko-Pastur isometry, at least at the beginning of the training.

This means that we are looking for activation functions whose Hermite expansion only includes terms for polynomials of order $H_2$ and above - so that $f$ is in the vector space $\mathcal{H} = Vec \left( H_{k}(x) ,\quad k \geq 2 \right)$. Zeroing out Hermite coefficients of order 0 and 1 is \emph{projecting} $f$ onto $\mathcal{H}$. Further, we also project on a sphere by enforcing $\eta = 1$ by scaling with $\gamma$.

In practice, this means that we can simply precompute the Gaussian integrals above as scalars $\alpha, \beta, \gamma$, and then replace any activation function $f$ by its \emph{Hermite normalization} $f_H$
\begin{equation}\label{sanbis}
f(x) \leftarrow f_{H}(x) = \frac{f(x) - \alpha x - \beta}{\gamma}
\end{equation}

Here we also note that 
\begin{equation}
\frac{d }{dx} f_{H}(x) = \frac{1}{\gamma} \left(\frac{d}{dx} f(x) - \alpha \right)
\end{equation}
which means that this transformation represents fixed centering and scaling of the backpropagation gradients passing through the activations. In this sense, it can be interpreted as a static version of batch normalization that is not backpropagated through (hence runs significantly faster).

\subsection{Normalization coefficients for various activation functions}
 Numerous activation functions are used in deep learning. If the ReLU unit is by now the most common, many others, such as the \emph{softplus} $x \rightarrow \ln(1+ \exp(x))$, exponential linear units, Gaussian exponential linear units, the \emph{swish} \citep{elfwing2018sigmoid} activation function, the sigmoid activation function, the hyperbolic tangent, and so on and so forth, are also present. We can apply static activation normalization (equation \ref{san}) to all of these and compute the $\alpha$, $\beta$ and $\gamma$ coefficients. The numerical results are presented in the table below.
 
\begin{figure}[h]\label{numcoeffs}
\begin{center}
 \begin{tabular}{||c c c c c c||} 
 \hline
 Activation & $\alpha$ & $\beta$ & $\gamma$ & $(f'^2 \mathcal{D}z)^2$ & $f'^4 \mathcal{D}z$ \\ [0.5ex] 
 \hline\hline
ReLU & 0.398942 &	0.5	& 0.301405	& 0.25	& 0.5 \\ 
 \hline
Softplus &	0.806059 & 0.5	& 0.146678	& 0.0680713	& 0.131594\\
 \hline
Sigmoid	& 0.5 &	0.206621	& 0.0262071	& 0.0680713	& 0.131594\\
 \hline
Tanh &	0	& 0.605706	& 0.165576	& 0.21567	& 0.341509\\
 \hline
GELU &	0.325735	& 0.5	& 0.323942	& 0.239622	& 0.497433\\  
 \hline
Swish &	0.206621	&0.5	&0.251164	&0.144007	&0.286581\\
 \hline
 ELU &	0.160521	&0.761578	&0.197932	&0.44636	&0.594411\\
 \hline
 x * Tanh(x)	& 0.605706&	0&	0.625308	&0.749437	&1.01452 \\[1ex]
 \hline
\end{tabular}
\end{center}
\label{fig:numcoeffs}
\caption{Numerical integral coefficients $\alpha$, $\beta$ and $\gamma$ for static activation normalization. Also included are the coefficients (higher order moments) related to the Gaussian kurtosis calculation $f'^4 \mathcal{D}$.}
\end{figure}

\subsection{Tilted ReLU}

In what follows we focus exclusively on the static normalization of the ReLU function, which gives a new activation where the associated constants can be derived in closed form. We get to a multiplicative constant factor (chosen to make its Lipschitz constant 1):
\begin{equation}
f_H(x) = |x|- \sqrt{\frac{2}{\pi}}
\end{equation}
In all our testing, this unit seems to yield the best results (excluding the normalized GELU which unfortunately yielded prohibitive computation times). Crucially, it is convex, just like the original ReLU unit. We thereafter refer to this unit as the \emph{tilted} ReLU, depicted below.

\begin{figure}[h]
\includegraphics[width=0.5\textwidth]{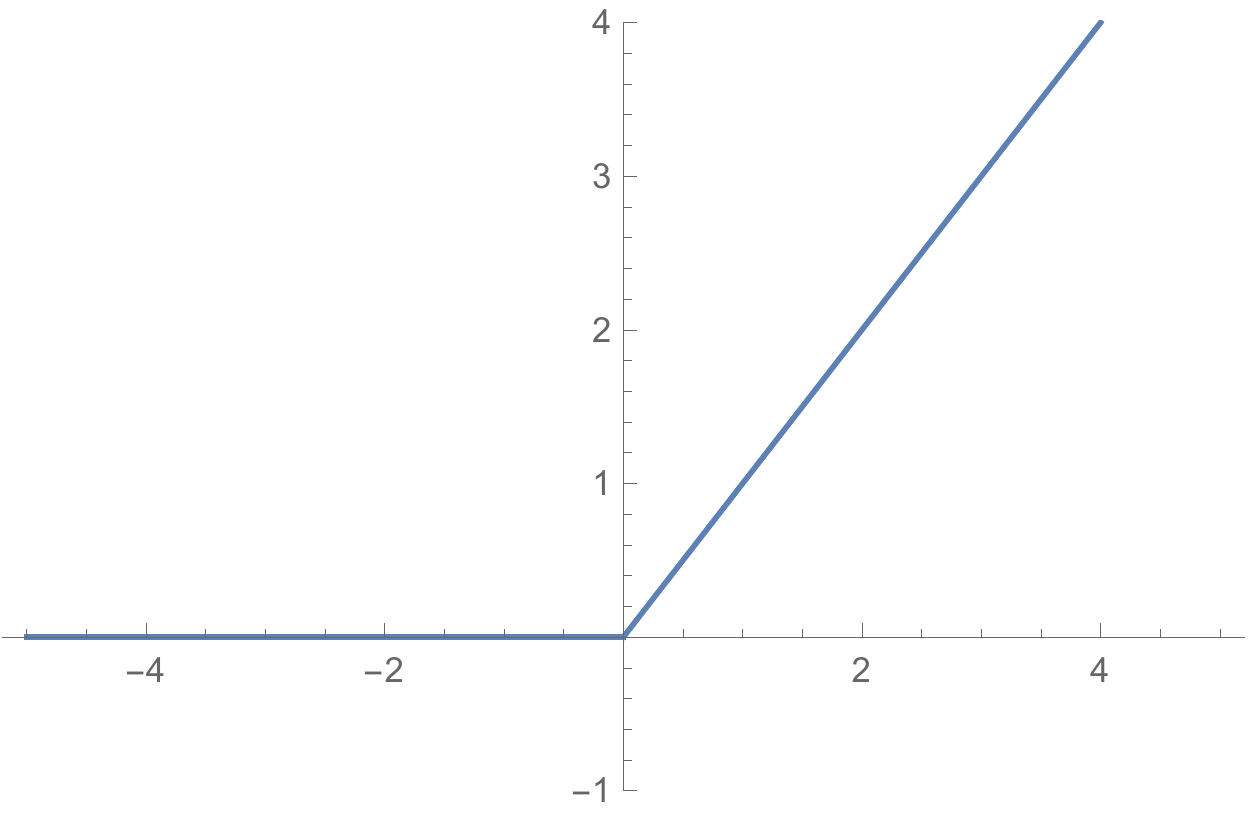}
\includegraphics[width=0.5\textwidth]{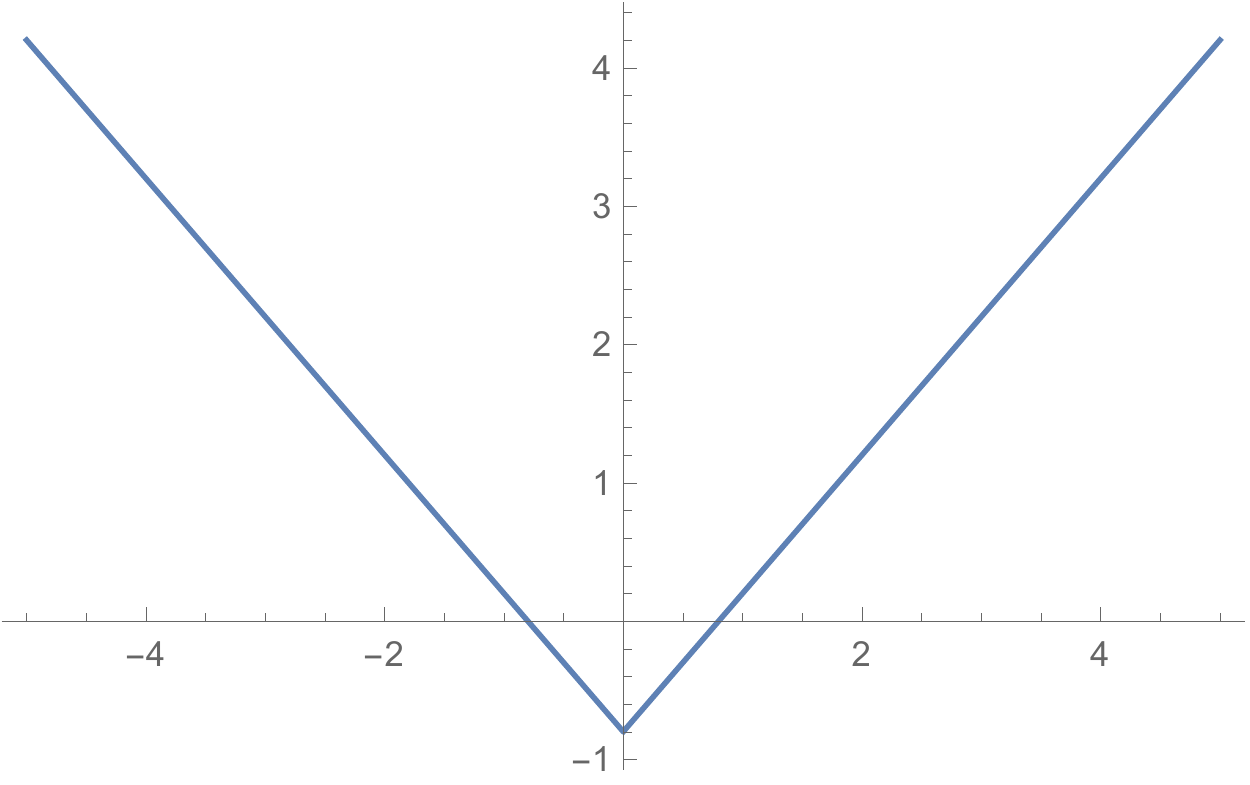}
\caption{The standard ReLU unit, as well as the \emph{tilted} ReLU unit (right) resulting from our activation normalization process applied to it.}
\label{fig:tiltedrelu}
\end{figure}

Interestingly, and while our process applies to any unit, the tilted ReLU can be seen as a special case of concatenated rectified linear units, which have previously shown to improve training \citep{crelu}. We now proceed to validate its use empirically.

\section{Experimental results}\label{sec:experiments}

\subsection{Setup}

Our experimental setup consists in training multi-layer perceptrons (feed-forward neural networks) on the CIFAR10 image classification dataset. The 32x32x3 image tensors are flattened into vectors of 3072 elements, and then fed as inputs to multi-layer perceptrons, consisting of a variable depth number of square layers, either 1024 or 128 units wide, and a final 10-way softmax layer. In general we compare 30-layer deep networks with networks of 60 layers or more. Batch normalization is not used by default, unless explicitly stated. Weight are initialized orthogonally \citep{saxe2013} by default. No loss function regularization is used. We use a mix of TensorFlow \citep{tensorflow} and Keras for our code, and run on a machine equipped with NVIDIA GTX1080 GPUs. We report on test-time accuracies, anytime performance, and robustness of convergence.
 
\subsection{Results}

\subsubsection{Convergence: trainability in layers}

We progressively increase the number of layers in our feedforward network until the training process fails. In this setup, the \emph{tilted ReLU}, even without batch normalization, provides convergence all the way to 60 layers, where all things considered the same setup with ReLUs stops converging after depth reaches circa 30 layers. This is of note for a simple affine tilting transformation, and provides simple validation for an involved theory.

\subsubsection{Convergence: speed, terminal accuracy, and impact of bias shift}

\begin{figure}[h]
\includegraphics[width=0.33\textwidth]{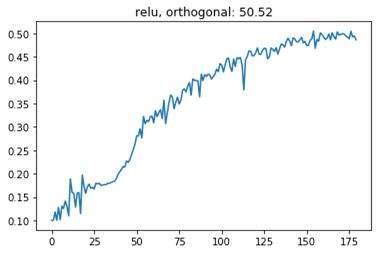}
\includegraphics[width=0.33\textwidth]{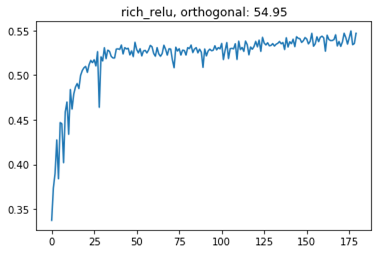}
\includegraphics[width=0.33\textwidth]{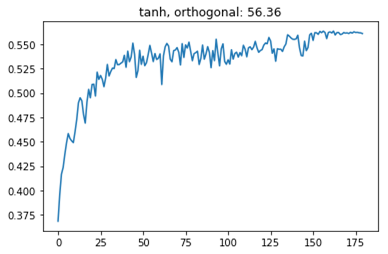}
\caption{Comparison of convergence for a 30 layer, 1024 width, feedforward network on CIFAR10. Test-time accuracy in percentage is plotted as a function of the number of training epochs. The ReLU unit barely converges, whereas the tilted ReLU (middle unit) brings accuracy and convergence speed in line with the hyperbolic tangent unit.}
\label{fig:figure3}
\end{figure}

Figure \ref{fig:figure3} shows our tilted ReLU unit, in the middle, shows terminal accuracy in line with the best performing unit for feedforward networks with orthogonal initialization, the hyperbolic tangent (rightmost). Furthermore, it gets there extremely fast, after about 25 epochs of training, in stark contrast with the standard ReLU unit which across many runs shows difficulty converging.

\begin{figure}[h]
\includegraphics[width=0.33\textwidth]{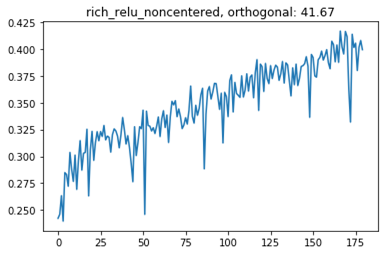}
\includegraphics[width=0.33\textwidth]{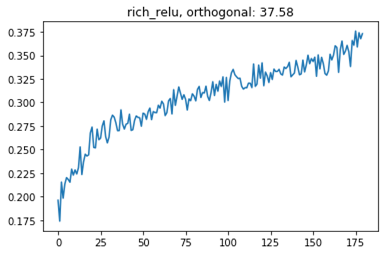}
\includegraphics[width=0.33\textwidth]{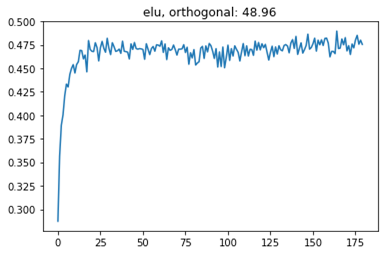}
\caption{Comparison of convergence for a 30 layer, 128 width, feedforward network on CIFAR10. The noncentered tilted unit, leftmost, performs as well (actually better) as the centered one. The decreased layer width still places us in the domain of validity of random matrix theory, but tilting is now much less efficient as judged by terminal accuracy, whereas the ELU unit doesn't suffer.}
\label{fig:figure4}
\end{figure}

Since tilting other activations doesn't seem to provide the same performance gains than on ReLUs, we question whether it's the tilting part or simply removing the shift part via substracting a constant (the $\beta = \langle f, 1 \rangle_\mathcal{D}$ term in equation \ref{sanbis}) that is responsible for improving the trainability of the network. We verify (see Figure \ref{fig:figure4})  that moving the tilted ReLU from $|x| - \sqrt{\frac{2}{\pi}}$ to $|x|$ still maintains convergence, showing that the tilting is the essential part of the improved convergence.

\subsection{Inspection of pre-activation eigenvalues}

In order to shed light on what is happening inside the networks, we decide to diagonalize $W \cdot W^T$ numerically, so as to look at the squared magnitude of the eigenvalues at each layer, $\rho_l(x)^2$. For convenience, we sort these by decreasing value.

Layer-wise, it appears that the spread of squared eigenvalue spectra decreases as we progress deeper into the layer. These spectra progressively flatten to converge to something close to the uniform 1 function.

\begin{figure}[h]
\includegraphics[width=0.5\textwidth]{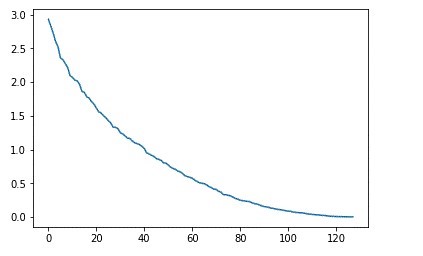}
\includegraphics[width=0.5\textwidth]{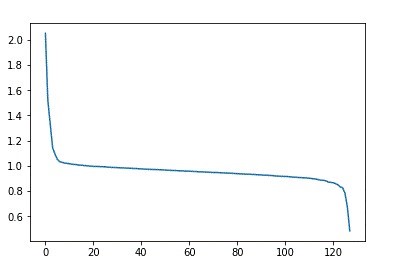}
\caption{Typical patterns for ordered, pre-activation eigenvalue moduli after training. Left : early layers of the neural network. Right : final layers. The spectrum clearly flattens with depth. With the \emph{tilted ReLU} unit, the spectrum, initialized at a constant of 1, remains very flat to full accuracy in a large part of the input vector universe. These plots remain in the 1024-width layer case.}
\label{fig:figure6}
\end{figure}

This has important potential implications for neural network training. It appears that in the orthogonal initialization case, the vast majority of squared eigenvalues is 1 or close to 1 (in magnitude) in the deepest layers of the network (see figure \ref{fig:figure6}). This is irrespective of the activation function used. In other words, there is a large subspace of data where the network layers are simply acting as a rotation operator. This could inform future work on optimization methods. At the same time, and in a manner compatible with building hierarchical features and representations of the input data, the right tail of the distribution tapers off to zero, so as to forget information. The left tail is most interesting as it looks skewed - seemingly, the exploding gradients problem is of more practical importance than the vanishing gradients one in our application.

Such a flat spectral distribution also provides a nice ex-post interpretation for \emph{residual networks}. These force the network training process to look for solutions near the identity matrix with a full flat spectrum. This way, the optimization process focuses on looking for distinctive features (the aforementioned tails). In light of our experimental results, this is particularly important deeper in the network.

In line with intuition, we also verify experimentally that batch normalization generates an eigenvalue spectrum even closer to the idealized Dirac distribution at one, with less vertical spread (but coming at computational expense).

\section{Conclusion and further work}\label{sec:conclusion}

We have defined static activation normalization as a projection of activation functions used in deep learning on a Hermite polynomial space. Empirically we observe that simply normalizing the ReLU unit in such a way enables us to train much deeper networks, and promotes robustness and convergence speed of the training process, making a first step towards 'pushing batch normalization within the activation'. Further work will focus on a better understanding of how to mix all the different competing constraints that come into play in activation design, possibly combining both static and dynamic isometry. Hopefully, that work will enable pro-active \emph{activation scoring} from such properties as their Gaussian moments. Moreover, the static isometry view also indicates that significant meaning could be found in the large number of pre-activation eigenvalues with moduli 1. Either constraining the optimization process to operate block-wise within the space of rotation matrices, or using random unitary matrix theory to give further predictions for the phase of eigenvalues and output vectors, could be fruitful research directions in the future.

\section{Acknowledgements}
The author wants to thank Bilal Piot, Mo Azar, Tiago Pereira, and Pratik Chaudhari for helpful discussions and comments.

\bibliography{iclr2019_conference}
\bibliographystyle{iclr2019_conference}

\end{document}